\documentclass[twoside,twocolumn,9pt]{article}
\usepackage{extsizes}
\usepackage[super,sort&compress,comma]{natbib} 
\usepackage[version=3]{mhchem}
\usepackage[left=1.5cm, right=1.5cm, top=1.785cm, bottom=2.0cm]{geometry}
\usepackage{balance}
\usepackage{mathptmx}
\usepackage{sectsty}
\usepackage{graphicx} 
\usepackage{lastpage}
\usepackage[format=plain,justification=justified,singlelinecheck=false,font={stretch=1.125,small,sf},labelfont=bf,labelsep=space]{caption}
\usepackage{float}
\usepackage{fancyhdr}
\usepackage{fnpos}
\usepackage[english]{babel}
\addto{\captionsenglish}{%
  
}
\usepackage{array}
\usepackage{droidsans}
\usepackage{charter}
\usepackage[T1]{fontenc}
\usepackage[usenames,dvipsnames]{xcolor}
\usepackage{setspace}
\usepackage[compact]{titlesec}
\usepackage{hyperref}

\usepackage[utf8]{inputenc} 
\usepackage[T1]{fontenc}    
\usepackage{hyperref}       
\usepackage{url}            
\usepackage{booktabs}       
\usepackage{amsfonts}       
\usepackage{nicefrac}       
\usepackage{microtype}      
\usepackage{xcolor}         
\usepackage{graphicx}
\usepackage{multirow}
\usepackage{longtable}
\usepackage{bm}
\usepackage{color}
\usepackage{makecell}
\usepackage{comment}
\usepackage{subfigure}
\usepackage{ulem}

\usepackage{epstopdf}

\definecolor{cream}{RGB}{222,217,201}

\begin{document}

\pagestyle{fancy}
\thispagestyle{plain}
\fancypagestyle{plain}{
\renewcommand{\headrulewidth}{0pt}
}

\makeFNbottom
\makeatletter
\renewcommand\LARGE{\@setfontsize\LARGE{15pt}{17}}
\renewcommand\Large{\@setfontsize\Large{12pt}{14}}
\renewcommand\large{\@setfontsize\large{10pt}{12}}
\renewcommand\footnotesize{\@setfontsize\footnotesize{7pt}{10}}
\makeatother

\renewcommand{\thefootnote}{\fnsymbol{footnote}}
\renewcommand\footnoterule{\vspace*{1pt}%
\color{cream}\hrule width 3.5in height 0.4pt \color{black}\vspace*{5pt}} 
\setcounter{secnumdepth}{5}

\makeatletter 
\renewcommand\@biblabel[1]{#1}            
\renewcommand\@makefntext[1]%
{\noindent\makebox[0pt][r]{\@thefnmark\,}#1}
\makeatother 
\renewcommand{\figurename}{\small{Fig.}~}
\sectionfont{\sffamily\Large}
\subsectionfont{\normalsize}
\subsubsectionfont{\bf}
\setstretch{1.125} 
\setlength{\skip\footins}{0.8cm}
\setlength{\footnotesep}{0.25cm}
\setlength{\jot}{10pt}
\titlespacing*{\section}{0pt}{4pt}{4pt}
\titlespacing*{\subsection}{0pt}{15pt}{1pt}

\fancyfoot{}
\fancyfoot[LO,RE]{\vspace{-7.1pt}\includegraphics[height=9pt]{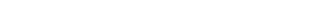}}
\fancyfoot[CO]{\vspace{-7.1pt}\hspace{13.2cm}\includegraphics{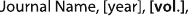}}
\fancyfoot[CE]{\vspace{-7.2pt}\hspace{-14.2cm}\includegraphics{head_foot/RF}}
\fancyfoot[RO]{\footnotesize{\sffamily{1--\pageref{LastPage} ~\textbar  \hspace{2pt}\thepage}}}
\fancyfoot[LE]{\footnotesize{\sffamily{\thepage~\textbar\hspace{3.45cm} 1--\pageref{LastPage}}}}
\fancyhead{}
\renewcommand{\headrulewidth}{0pt} 
\renewcommand{\footrulewidth}{0pt}
\setlength{\arrayrulewidth}{1pt}
\setlength{\columnsep}{6.5mm}
\setlength\bibsep{1pt}

\makeatletter 
\newlength{\figrulesep} 
\setlength{\figrulesep}{0.5\textfloatsep} 

\newcommand{\topfigrule}{\vspace*{-1pt}%
\noindent{\color{cream}\rule[-\figrulesep]{\columnwidth}{1.5pt}} }

\newcommand{\botfigrule}{\vspace*{-2pt}%
\noindent{\color{cream}\rule[\figrulesep]{\columnwidth}{1.5pt}} }

\newcommand{\dblfigrule}{\vspace*{-1pt}%
\noindent{\color{cream}\rule[-\figrulesep]{\textwidth}{1.5pt}} }

\makeatother

\twocolumn[
  \begin{@twocolumnfalse}
{\includegraphics[height=30pt]{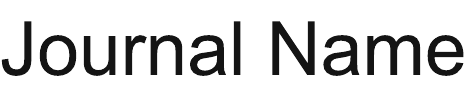}\hfill\raisebox{0pt}[0pt][0pt]{\includegraphics[height=55pt]{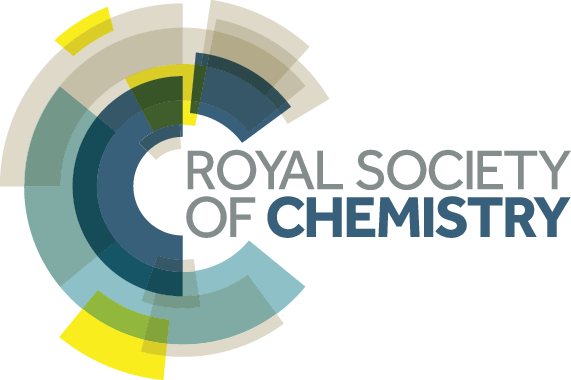}}\\[1ex]
\includegraphics[width=18.5cm]{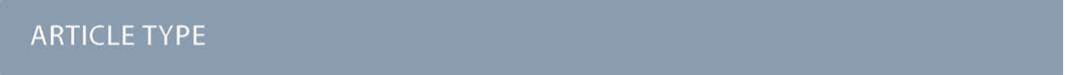}}\par
\vspace{1em}
\sffamily
\begin{tabular}{m{4.5cm} p{13.5cm} }

\includegraphics{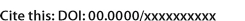} & \noindent\LARGE{\textbf{From Tokens to Materials: Leveraging Language Models for Scientific Discovery}} \\
\vspace{0.3cm} & \vspace{0.3cm} \\

 & \noindent\large{Yuwei Wan$^{b,c,\ddag}$, Tong Xie$^{a,b,\ddag,\ast}$,Nan Wu$^{c}$, Wenjie Zhang$^{d}$, Chunyu Kit$^{c}$, Bram Hoex$^{a,\ast}$}
 \\

\includegraphics{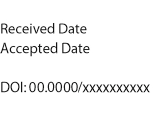} & \noindent\normalsize{Exploring the predictive capabilities of language models in material science is an ongoing interest. This study investigates the application of language model embeddings to enhance material property prediction in materials science. By evaluating various contextual embedding methods and pre-trained models, including Bidirectional Encoder Representations from Transformers (BERT) and Generative Pre-trained Transformers (GPT), we demonstrate that domain-specific models, particularly MatBERT significantly outperform general-purpose models in extracting implicit knowledge from compound names and material properties. Our findings reveal that information-dense embeddings from the third layer of MatBERT, combined with a context-averaging approach, offer the most effective method for capturing material-property relationships from the scientific literature. We also identify a crucial "tokenizer effect," highlighting the importance of specialized text processing techniques that preserve complete compound names while maintaining consistent token counts. These insights underscore the value of domain-specific training and tokenization in materials science applications and offer a promising pathway for accelerating the discovery and development of new materials through AI-driven approaches.
} \\

\end{tabular}

 \end{@twocolumnfalse} \vspace{0.6cm}

  ]

\renewcommand*\rmdefault{bch}\normalfont\upshape
\rmfamily
\section*{}
\vspace{-1cm}





\footnotetext{\textit{$^{\ddag}$~Equal contribution}}
\footnotetext{\textit{$^{\ast}$~Corresponding author, tong@greendynamics.com.au, b.hoex@unsw.edu.au}}
\footnotetext{\textit{$^{a}$~School of Photovoltaic and Renewable Energy Engineering, University of New South Wales, Kensington, NSW, Australia}}
\footnotetext{\textit{$^{b}$~GreenDynamics, Kensington, NSW, Australia.}}
\footnotetext{\textit{$^{c}$~Department of Linguistics and Translation, City University of Hong Kong, Hong Kong, China}}
\footnotetext{\textit{$^{d}$~School of Computer Science and Engineering, University of New South Wales, Kensington, NSW, Australia}}






\section{Introduction}

Materials science, an inherently versatile field, intersects with physics, chemistry, energy studies, and engineering, driving innovation across numerous industries. This interdisciplinary nature has led to groundbreaking discoveries such as graphene\cite{geim2009graphene}, high-temperature superconductors\cite{blatter1994vortices}, and metal-organic frameworks\cite{furukawa2013chemistry}. However, it also presents a significant challenge: efficiently identifying materials with optimal properties for specific purposes from vast literature findings. Traditionally, advances have relied on time-consuming and resource-intensive trial-and-error experimentation guided by researchers' expertise.\cite{yan2024machine} The pressing need to address global challenges demands a more rapid, efficient, and cost-effective approach to materials discovery.

Recent works \cite{tshitoyan2019unsupervised, shetty2021automated} have utilized unsupervised natural language processing (NLP) techniques, particularly word embeddings, to discover new materials and synthesize materials science knowledge. These techniques represent words as vectors in a high-dimensional space, capturing complex materials science concepts and structure-property relationships directly from text without explicit domain knowledge insertion \cite{pilania2021machine}. While word embeddings trained by algorithms like Word2Vec \cite{mikolov2013efficient} and GloVe \cite{pennington2014glove} have gained popularity, more advanced models like Bidirectional Encoder Representations from Transformers (BERT) \cite{devlin2018bert} and Generative Pre-trained Transformers (GPT) \cite{brown2020language} have emerged, capable of capturing even richer contextual information. BERT and GPT utilize attention mechanisms and transformer architectures, allowing them to consider the entire context of a word in a sentence rather than just a fixed window of surrounding words \cite{brown2020language}. This enables them to capture long-range dependencies and nuanced relationships between words, addressing limitations of earlier models in handling polysemy (words with multiple meanings) and context-dependent sentiment \cite{devlin2018bert}. For instance, while Word2Vec and GloVe struggle with words like "bank" (financial institution or river bank) or context-dependent sentiments, BERT and GPT can distinguish between these based on the surrounding context, potentially leading to more accurate representations of materials science concepts \cite{tshitoyan2019unsupervised}.

This study assesses the feasibility of using contextualized word representations and semantically rich sentence embeddings for material prediction, aiming to accelerate scientific discovery in materials science. We investigated various contextual embedding models, including pre-trained BERT and GPT models, on a thermoelectric material ranking prediction task. Our findings emphasize the importance of domain-specific pretraining and tokenization techniques, and identify a "tokenizer effect" in embedding models which uses subword tokenization mechanism. By analyzing relationship between token length and prediction performance, we find that breaking down compositions in materials science into too many tokens will lead to a loss of contextual information, which limits the performance of language models in material prediction. To address this problem, we developed a sentence embedding model SentMatBERT\_MNR by combining a materials-specific BERT model with a pooling layer and fine-tuning with natural language inference triplets and material description pairs in a contrastive learning framework. We used a tokenizer pretrained on materials science text to mitigate the tokenizer effect. And we propose a context-average approach to enhance model performance by providing better contextual understanding of material names through sentences containing these names, enabling more efficient knowledge assimilation. SentMatBERT\_MNR with contextual inputs achieves strong correlation 0.5919 between prediction and experimental results, surpassing Word2Vec method \cite{tshitoyan2019unsupervised} by 7 points and DFT baseline \cite{tshitoyan2019unsupervised} by 28 points.

Our method can quickly process and contextualize information from the vast and ever-growing body of scientific literature, reducing the time required for literature reviews and hypothesis generation. Moreover, by identifying patterns in existing materials data, our method has the potential to reduce the need for intuition-driven experimentation and extensive human trial-and-error, which traditionally consume considerable time and resources. Unlike previous static word embedding methods in materials science, which cannot capture nuanced contextual information, our models provide richer representations of materials and their properties, enabling more accurate predictions. Compared with large language models (LLMs), our lightweight model has domain-specific training experience and large amount of contextual inputs, surpassing both zero-shot and few-shot results of ChatGPT, while with much higher computational efficiency. These advancements could lead to faster identification of promising materials for specific applications. The study represents a pioneering effort to accelerate materials discovery by leveraging advanced language models and experimental knowledge embedded in literature.

To encourage further exploration, we have made our code and datasets publicly available at https://github.com/MasterAI-EAM/Matter2Vec.

\section{Related works}

Knowledge discovery was defined as 'the non-trivial process of identifying valid, novel, potentially useful and ultimately understandable pattern or knowledge in data' in 1996 \cite{fayyad1996knowledge}. Word embedding technique, which has its root in distributional semantics \cite{firth1957synopsis}, can be applied in knowledge discovery prediction directly \cite{aceves2023mobilizing}. It often rely on semantic-similarity measures between word representations to predict relationships, which are subsequently validated using domain-specific scientific methods \cite{panesar2022biomedical,chen2024prediction}. Traditional approaches have leveraged static word embeddings (each word has fixed vector) such as Word2Vec and GloVe to uncover latent knowledge within domain-specific text data. For example, Tshitoyan et al. \cite{tshitoyan2019unsupervised} proposed an embedding-based ranking prediction: Word2Vec embeddings of material names were tanked by their cosine similarity to the embedding of 'thermoelectric' and obtained a 52\% rank correlation with experimental zT (thermoelectrical property) results. In contrast, the ranking predicted by density functional theory (DFT) calculation using power factor (another related thermoelectrical property) only exhibits a 31\% rank correlation. The context analysis reveals that the direct relationship between the novel materials and ‘thermoelectric’ may be attributed to indirect connections involving the material names and related terms, such as 'chalcogenide' (many chalcogenides are good thermoelectrics) and 'band gap' (important to thermoelectric properties). This method was applied to solar materials and identified potential candidates such as As2O5 \cite{zhang2022unsupervised}. Shetty and Ramprasad also demonstrated that word embeddings trained on a corpus of polymer papers could encode materials science knowledge and used to identify novel polymers for certain applications \cite{shetty2021automated}. In biomedical domain, Venkatakrishnan et al. \cite{venkatakrishnan2020knowledge} applied the same method to identify novel tissue-reservoirs of the ACE2 receptor used by SARS-CoV-2 to invade a host. 

Several of the above studies \cite{tshitoyan2019unsupervised,venkatakrishnan2020knowledge} suggested in the discussion that dynamic embeddings (each word is assigned a distinct embedding based on its context) from pre-trained models like BERT might surpass those static embeddings in the realm of knowledge discovery. In a sentiment analysis study based on Twitter \cite{deb2022comparative}, dynamics embeddings demonstrate a better ability to capture hidden information and intricate relationships from context than static embeddings. However, there are few efforts to tailor models like BERT for knowledge discovery prediction purposes. Panesar \cite{panesar2022biomedical} obtained word embeddings based on sentences extracted from a biomedical corpus using domain-specific pretrained models like BioBERT \cite{lee2020biobert}, achieving superior performance on biomedical benchmarks compared to static embeddings. The benchmarks utilized in the paper involves directly comparing the cosine similarity of word representations from language models to human domain-expert ratings. This kind of task aims to measure ability of capturing semantic relationship, which differs from material prediction tasks based on experimental results, leaving room for further exploration.

Reimers and Gurevych \cite{reimers2019sentencebert} presented a modification of the pre-trained BERT network Sentence-BERT (SBERT), which was finetuned with Siamese and Triplet network structures to derive semantically meaningful sentence embeddings. While Siamese network takes two inputs, Triplet network takes three inputs – an anchor sample, a positive sample similar to the anchor, a negative sample dissimilar to the anchor, and learns an embedding space during training where the anchor is closer to the positive sample than the negative sample based on a distance metric. Trained on the combination of SNLI \cite{bowman2015large} and MultiNLI \cite{williams2018broadcoverage} datasets, SBERT outperformed other sentence embedding methods significantly over Semantic Textual Similarity (STS) tasks. Lamsal et al. \cite{Lamsal2024CrisisTransformers} stated that it is critical to have semantically meaningful embeddings that position sentences in a vector space such that semantically similar sentences are located closely together for effective semantic search and clustering tasks. Their domain-specific pre-trained model CrisisTransformer was further fine-tuned through contrastive learning objectives with sentence pairs from GooQA \cite{GooAQ} and sentence triplets from AllNLI \cite{bowman2015large,williams2018broadcoverage} and QQP, and mean pooling over the token embeddings with attention was implemented to generate sentence embeddings. Gao et al. \cite{SimCSE} proposed a contrastive learning framework (SimCSE) for producing sentence embeddings with natural language inference datasets, using entailments as positive samples and contradiction as negative samples.

\section{Methods}

BERT and GPT models generate distinct vectors for the same word in varying contexts, based on the surrounding words within the sentence. We employed various approaches to acquire contextual embeddings for material prediction task, encompassing methodologies involving the exclusion or inclusion of actual sentence contexts, different layers, and combinations of layers. This section provides a detailed exposition of two methods for acquiring word embeddings of material names from BERT and MatBERT in our study, namely, the context-free and context-average.

\subsection{Context-free} 
\label{sec:context-free}
\begin{figure*}[h!]
 \centering
 \subfigure[context-free]{
 \includegraphics[width=0.6\textwidth]{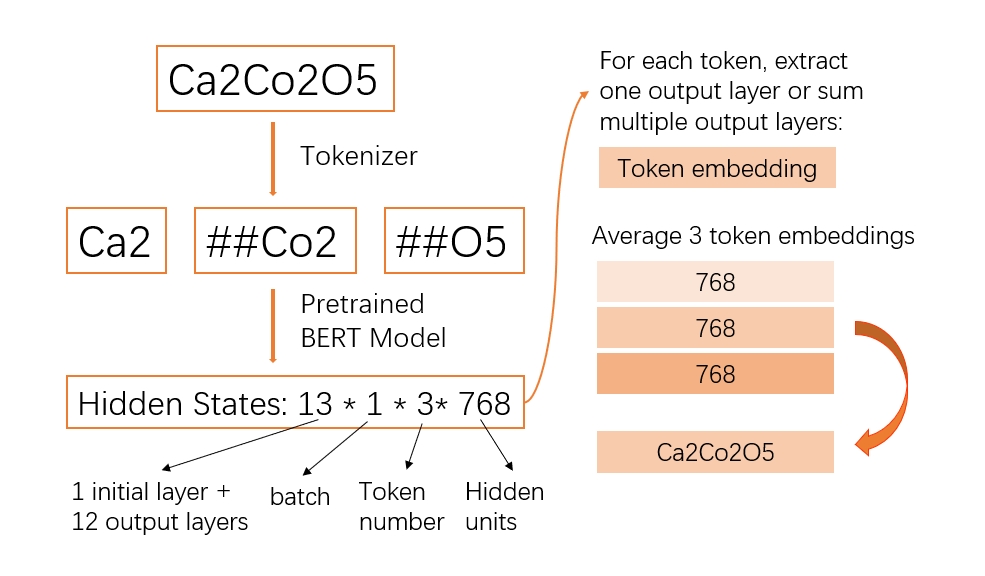}
  \label{fig:context-free}
 }
 \subfigure[context-average]{
 \includegraphics[width=0.7\textwidth]{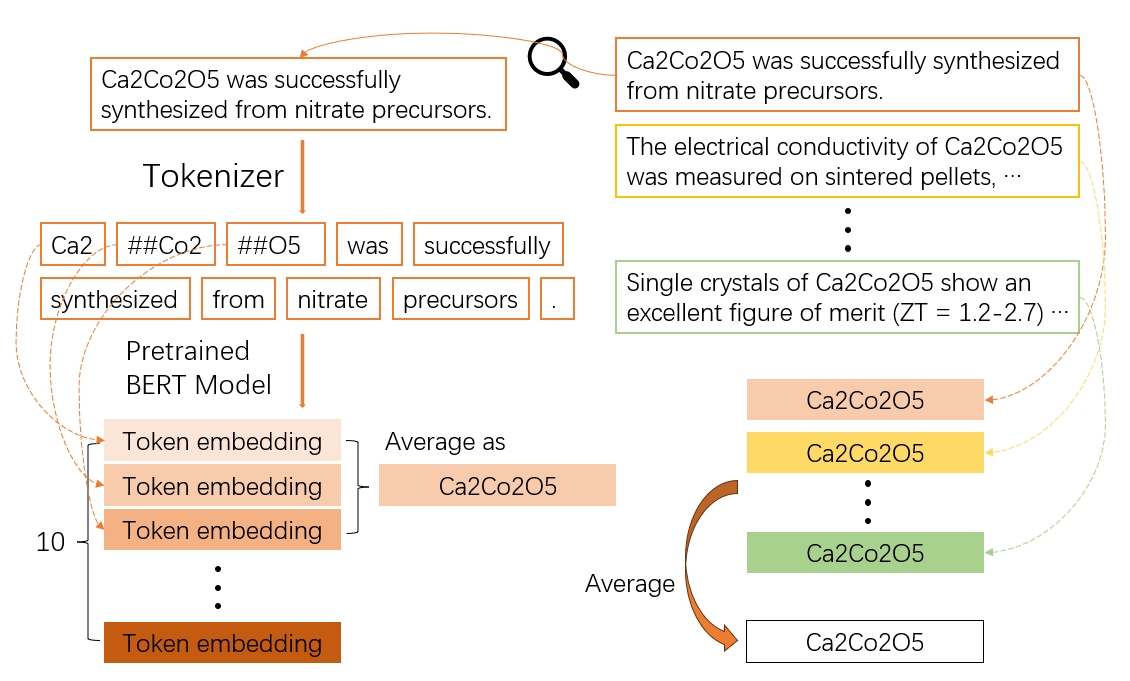}
 \label{fig:context-average} 
 }
 
 \caption{A diagram of obtaining context-free and context-average embedding for Ca2Co2O5. For both context-free and context-average methods, material names are segmented into tokens and input to the token embedding model BERT. Taking the vectors from certain layer of the model's output as token embeddings, the embedding of a material name is the average of the embeddings of its tokens. The difference between these two methods is that, in context-average method, material names are accompanied by several context sentences, and the final material representation is the average over material embeddings regarding different context sentences.}
\end{figure*}

We use a single context \cite{bommasani2020interpreting} for each material name to obtain context-free BERT word embebddings. Initially, we utilized the pretrained BERT tokenizer to convert material names into tokens. For each material name word $w$, tokenizer decomposed $w$ into $x$ sub-word tokens ($x\geq1$), resulting in tokenized text $\{t_1, ..., t_x\}$. Subsequently, we fed the tokenized text into the pretrained BERT model, resulting in output hidden states with four dimensions: 
\begin{enumerate}
  \item The layer number (13, including the initial layer and 12 output layers)
  \item The batch number (always 1 corresponding to a single input)
  \item The token number (varying as number of tokens in the input)
  \item The hidden unit number (768 features)
\end{enumerate}
As Figure \ref{fig:context-free} shows, each token $t$ within our input generated 13 separate vectors, each of which possessed a length of 768. To form a token embedding ${t}$, we directly extracted certain layer vector or combine some of the layer vectors by operations like summing, averaging, or concatenating. Given that there were no contextual words available to impart meaning to the material name in the input, we calculated the final representation by averaging across all token embeddings: ${w}=\frac{1}{x}({t}_1+...+{t}_x)$. 

\subsection{Context-average} 
\label{sec:context-average}
Inspired by Bommassani et al. \cite{bommasani2020interpreting}, we also aggregated contextualized word representations over contexts to obtain static-equivalents. Utilizing a paper downloading tool \href{https://github.com/MasterAI-EAM/SciCrawler}{(https://github.com/MasterAI-EAM/SciCrawler)}, we collected scholarly literature related to the material names from Web of Science (English; published in 2000.01.01-2023.09.15; published by Elsevier or Springer Nature) and converted them into plain texts. As Figure \ref{fig:context-average} shows, for each material name $w$, we select $n$ $(n\leq 100)$ sentences containing it as contexts $\{s_1, ..., s_n\}$ (if the number of collected sentences is larger than 100, randomly sample 100 sentences from the collection). These sentences were then tokenized by pretrained BERT tokenizer. Assuming that the tokens of $w$ (if $w$ have more than 1 occurrences in $s$, only use the first one) are located within the tokens of $s$ at the positions $[s_i, ..., s_j]$. For each sentence $s$, we obtained token embeddings for each token using same method of generateing token embeddings in context-free approach and extracted $\{\bm{t}_{s_i}, ..., \bm{t}_{s_j}\}$ from all token embeddings of $s$. The representation of $w$ in the context $s$ is calculated as:
\begin{equation}
    {w_s}=\frac{1}{j-i+1}({t}_{s_i}+...+{t}_{s_j})
\end{equation}
After obtaining representation of $w$ in all contexts, we calculated the arithmetic mean of $n$ contextual representations:
\begin{equation}
    {w}=\frac{1}{n}({w}_{s_1}+...+{w}_{s_n})
\end{equation}

For both approaches, we also obtained the representation of a pre-selected word about a certain application, like ’thermoelectric’, to be the center embedding. The material names were ranked by cosine similarity between their embeddings and the center embedding.

\section{Experiments}

\subsection{Datasets}
In order to assess the performance of contextual embeddings obtained through different methods and configurations in the context of a material prediction task, we conducted an initial exploration on an existing dataset. Tshitoyan et al. \cite{tshitoyan2019unsupervised} constructed a small-scale dataset using 84 materials that appear both in their text corpus and the experimental set \cite{gaultois2013data}. We reproduced a 84-material dataset ranked by zT value (an important component of the overall thermoelectric figure of merit) by using their released data. And as described in \ref{sec:context-average}, we collected context sentences for these materials from scientific literature.

For fine-tuning MatBERT model, we used AllNLI dataset \href{https://huggingface.co/datasets/sentence-transformers/all-nli}{(https://huggingface.co/datasets/sentence-transformers/all-nli)} and Quora Question Pairs (QQP) dataset \cite{quora-question-pairs}. AllNLI is a concatenation of the SNLI \cite{bowman-etal-2015-large} and MultiNLI \cite{N18-1101} datasets, containing anchor-entailment-contradiction triplets, while QQP dataset contains anchor-positive-negative triplets (see both type of triplets example in supplementary). All triplets indicate semantic similarity and dissimilarity, therefore are suitable for the first fine-tuning step which aims to improve model’s ability for recognizing textual similarity. These two datasets are commonly used together in contrastive learning frameworks to enhance the overall quality of embeddings \cite{ahmad2018learning,Lamsal2024CrisisTransformers}.

Our material description dataset used in fine-tuning were collected from two material glossary pages \href{https://en.wikipedia.org/wiki/Glossary_of_chemical_formulae}{(https://en.wikipedia.org/wiki/Glossary\_of\_chemical\_formulae)} and \href{https://en.wikipedia.org/wiki/List_of_inorganic_compounds}{(https://en.wikipedia.org/wiki/List\_of\_inorganic\_compounds)} by using English Wikipedia API. For each material, its chemical formula, English synonym, and a paragraph of description are extracted. The three components of each material forms similarity pairs between every two of them. Similarity pairs derived from this dataset are used in the second finetuning step which aims to cater the model with domain specific knowledge.

\subsection{Models}
For our thermoelectric material prediction task, we evaluated the performance of the following models:
\begin{itemize}
    \item \textbf{BERT} \cite{devlin2018bert} was pretrained on English language using a masked language modeling (MLM) and next sentence prediction (NSP) objective. In this study, we used a case-sensitive BERT, \href{https://huggingface.co/bert-base-cased}{bert-base-cased}. The texts used for training were tokenized using WordPiece \cite{schuster2012japanese} and reserved a vocabulary size of 28,996. 
    \item \textbf{MatBERT} \cite{trewartha2022quantifying} is a BERT model trained using scientific papers specifically from the field of materials science and only MLM objective. We used MatBERT-base-cased (available at \href{https://github.com/lbnlp/MatBERT}{https://github.com/lbnlp/MatBERT}). The texts used for training were also tokenized using WordPiece and reserved a vocabulary size of 30,552.
    \item \textbf{SentMatBERT} consists of MatBERT and a mean pooling layer without any finetuning, producing  sentence embeddings of 768 dimensions.
    \item \textbf{SentMatBERT\_MNR} is SentMatBERT fine-tuned on AllNLI, QQP and material description dataset with Multiple Negatives Ranking (MNR) \cite{henderson2017efficient} loss. MNR loss measures the difference between positive and negative samples concerning a query (see details in \ref{sec:finetune}). 
    \item \textbf{OpenAI Embeddings API} \cite{neelakantan2022text} can be directly used to directly retrieve contextual embeddings for text data. We used text-embedding-ada-002 in this study.
\end{itemize}
In addition, we also use ChatGPT (GPT-3.5, 28 October 2023) \cite{chatgpt}to re-rank the given materials as a supplementary trial. This chat mode does not involve embedding. Rather, it relies on interactive engagement with the model through prompts, enabling the model to re-rank the material using its inherent capabilities. The prompts used are available in supplementary.

\subsection{Fine-tuning of sentence embedding model SentMatBERT}
\label{sec:finetune}
MNR loss for two inputs (anchor, positive) maximizes the cosine similarity between an anchor sentence and its positive sentence while considering all other positive sentences in the batch as dissimilar, which are mapped further apart in the vector space. Given a batch of sentence pairs $(a_1, a_1^+)$, $(a_2, a_2^+)$, ..., $(a_n, a_n^+)$, it is assumed that $(a_i, a_i^+)$ are semantically similar while $(a_i, a_j^+), \forall j \ne i $ are semantically dissimilar. The training objective for $(a_i, a_i^+)$ within a batch of size N is:  
\begin{equation}
    loss_i = - log \frac{e^{cos\_sim(s_i,s_i^+)/\tau}}{\sum_{j=1}^{N} e^{cos\_sim(s_i,s_j^+)/\tau}}
\end{equation}
MNR for three inputs (anchor, positive, negative) considers the third sentence in a triplet as “hard negative” in addition to “in-batch negatives”. MNR loss generally achieves better result when provided with hard negatives. The training objective for triplet $(a_i, a_i^+, a_i^-)$ within a batch of size N is: \begin{equation}
    loss_i = - log \frac{e^{cos\_sim(s_i,s_i^+)/\tau}}{\sum_{j=1}^{N} ({e^{cos\_sim(s_i,s_j^+)/\tau} + e^{cos\_sim(s_i,s_j^-)/\tau}})}
\end{equation}
where $s_i$ and $s_i^+$ are sentence embeddings of $a_i$ and $a_i^+$ generated by MatBERT with a pooling layer, and $ \tau $ is the temperature parameter.

In the first finetuning step, SentMatBERT is trained with 370k sentence triplets from AllNLI and QQP and the corresponding MNR loss for three inputs. In the second finetuning step, SentMatBERT\_NLI is trained with 20k pairs from our material description dataset with MNR loss for two inputs. This two-step fine-tuning strategy adheres to the principles of domain adaptation \cite{xu2021gradual}, where a model is first trained on a general-domain dataset to capture broader knowledge, and then fine-tuned with domain-specific data to refine and specialize its understanding. The two-phase training design helps mitigate potential issues associated with combining multiple datasets by gradually transitioning from general to specialized knowledge, allowing the model to more effectively align with the material science domain. Parameters implemented can be found in Table 2. We used validation set for selecting the best epoch in the first-step fine-tuning. We selected the model from epoch 5 (out of 10 epochs) based on performance on the STSb dataset \cite{cer2017semeval} for hyperparameter optimization. We did not introduce validation set for second-step fine-tuning because there is lack of such benchmarks for domain-specific textual similarity, and we directly used epoch 5.

\begin{table}[h]
\small
  \caption{\ Parameters of SentMatBERT finetuning. MNR loss for three inputs uses triplets (anchor, positive, negative), while for two inputs it uses pairs (anchor and positive). In the first fine-tuning step, with many short artificial sentences, we use a larger batch size (256) and shorter max sequence length (75). The second fine-tuning step, with fewer and longer knowledge-dense descriptive texts, uses adjusted hyperparameters.}
  \label{tbl:example1}
  \begin{tabular*}{0.48\textwidth}{@{\extracolsep{\fill}}lll}
    \hline
    Parameters & First-step & Second-step \\
    \hline
    Loss & MNR loss for three inputs & MNR loss for two inputs \\
    Batch Size & 256 & 32 \\
    Max seq length & 75 & 128 \\
    Epochs & 5 & 5 \\
    Optimizer & AdamW & AdamW \\
    \hline
  \end{tabular*}
\end{table}

\section{Results}

To assess the performance of contextual embeddings, we compared the predicted rank with actual rank (ranked by the experimental results) using the Spearman's rank correlation coefficient \cite{spearman1904proof}. Spearman's correlation ranges from -1 to 1 (values near 1 indicate similarity in two ranks, while values near -1 indicate dissimilarity in two ranks). The baseline of this 84-material dataset should be density functional theory (DFT) method with around 31\% rank correlation \cite{tshitoyan2019unsupervised}. Their DFT calculations used  projector augmented wave (PAW) \cite{kresse1999ultrasoft} pseudopotentials and the Perdew–Burke–Ernzerhof (PBE) \cite{perdew1996generalized} generalized-gradient approximation (GGA), implemented in the Vienna Ab initio Simulation Package (VASP) \cite{kresse1996efficiency}. 
In our experiments with word embedding, we encountered results that deviated somewhat from our initial expectations. However, we identified certain patterns that align with phenomena previously observed in related studies and we observed intriguing phenomena associated with the tokenizers of pretrained models.

\subsection{Tokenizer effect in context-free word embedding}
As mentioned in context-free method in section \ref{sec:context-free}, we directly tokenized the 84 material names and input them into both the BERT and MatBERT models. For the sake of convenience in subsequent discussions, we shall denote the Spearman's correlation between the predicted rank of 84 materials and their actual rank as 'Correlation\_1', and the Spearman's correlation between the predicted rank of 84 materials and their tokenized length as 'Correlation\_2'. We selected 'thermoelectric' as the center word. We calculated Correlation\_1 and Correlation\_2 using embeddings generated by each layers in two models (Table \ref{tab:context-free} shows part of the results). The results indicate that the BERT embeddings generated using the context-free method do not exhibit a meaningful correlation with the predicted and actual results. However, the output of the tokenizer does influence the ranking to some extent, suggesting that material names composed of fewer tokens tend to be ranked higher.

In contrast, embeddings obtained from the domain-specific pretrained model MatBERT demonstrate a very subtle correlation between the predicted and actual results. Comparing the results across layers (excluding the initial layer), we observe that the correlation values peak at the third layer and rapidly decline, with a resurgence towards the end. Additionally, it is notable that the impact of the MatBERT tokenizer is more pronounced than that of BERT. Interestingly, the third layer, which exhibits the weakest influence from the tokenizer, yields the most favorable results.

\begin{figure}[h!]
 \centering
 \includegraphics[width=0.5\textwidth]{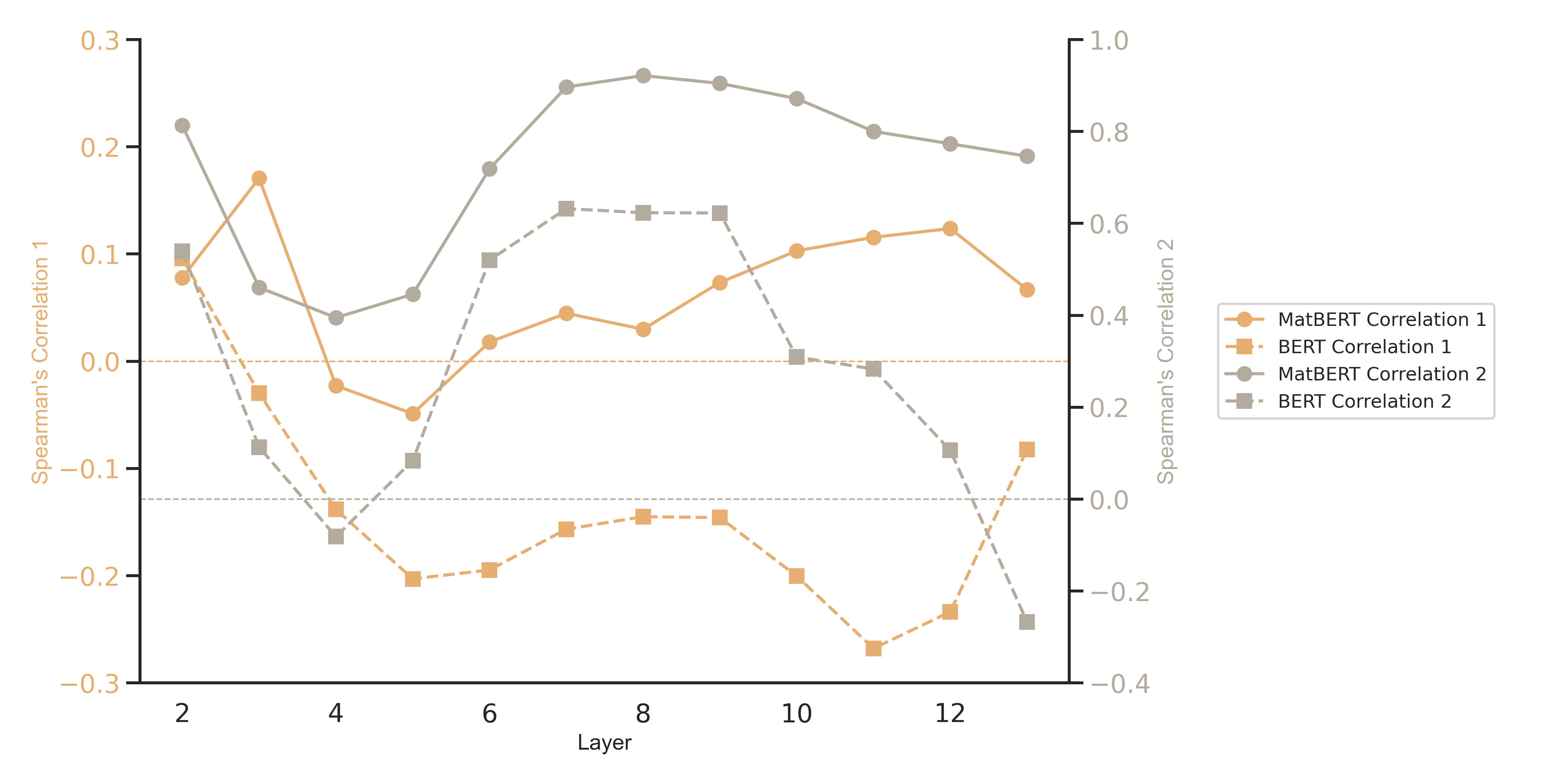}
 \caption{Results of thermoelectrical material prediction using context-free method. Correlation\_1 refers to the Spearman correlation between predicted ranking and standard ranking of the 84 materials in our evaluation dataset, reflecting the performance of pre-trained token embedding models – BERT and MatBERT, on the task of applicational material prediction. Correlation\_2 refers to the Spearman correlation between the predicted ranking and the ranked sequence of material token lengths, suggesting that the tokenized length of material names is an influential factor over task outcome.}
 \label{fig:context-avg-results}
\end{figure}

Regarding the impact of the tokenizer, we conducted several minor tests. First, we compare the length of tokenized 84 materials (see Fig.S1 in supplementary). Compared to BERT, MatBERT segments a material's name more comprehensively, without being overly fragmented. Then we calculated the Spearman's correlation between the actual rank of 84 materials and their tokenized length, which is only 1.6\%. This indicates that the shorter tokenized length cannot imply better experimental performance. We also replaced the center word "thermoelectric" with unrelated words like "apple" or "hit." After this substitution, both BERT and MatBERT failed to demonstrate even the slight correlation between the predicted rank and the actual rank. Notably, the tokenizer effect at lower layers significantly diminished (for instance, MatBERT exhibited a Correlation\_2 of -15\% at the third layer). However, at upper layers, Correlation\_2 remained relatively high, with both MatBERT and BERT reaching around 70\%.

\subsection{Tokenizer effect in context-average word embedding}
\begin{figure}[h!]
 \centering
 \includegraphics[width=0.5\textwidth]{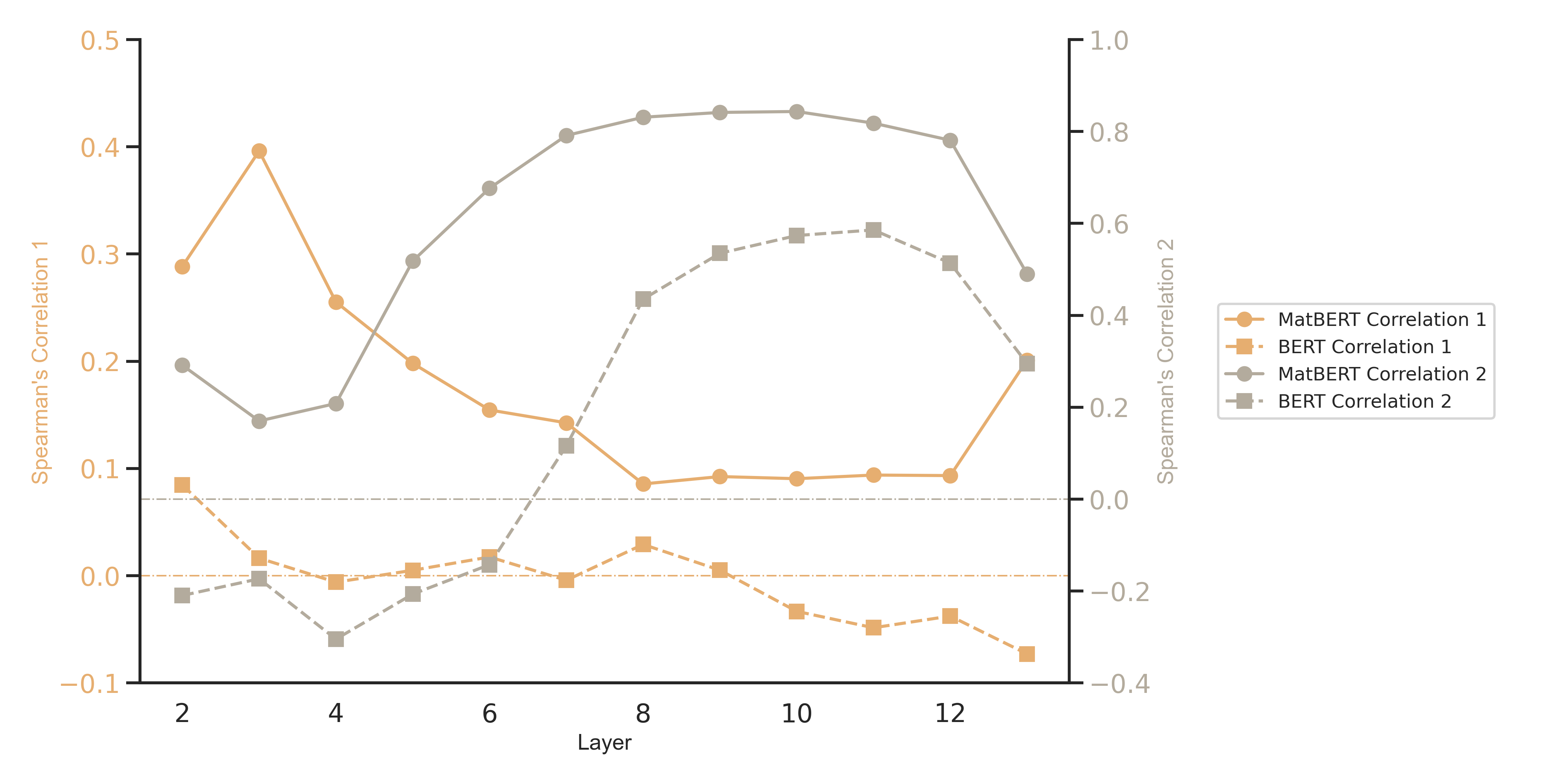}
 \caption{Results of thermoelectrical material prediction using context-average method. As pre-trained token embedding models output 13 layers of hidden states, each layer of high-dimensional vectors are extracted and employed as token embeddings separately. The layer from which token embeddings are extracted demonstrates impact over task result, and certain layer shows significant advantage over other layers.}
 \label{fig:context-avg-results}
\end{figure}
As mentioned in context-average method in section \ref{sec:context-free}, we directly tokenized the context sentences of 79 material names (5 material names were not found in web of science with predefined conditions) and input them into both the BERT and MatBERT models to generate embeddings of material names. As shown in Figure \ref{fig:context-avg-results}, the performance of context-average BERT embeddings aligns with that of BERT when utilizing the context-free method, but MatBERT shows a substantial improvement. For BERT, even in the case of the best-performing layer, the Correlation\_1 value is insufficient to establish a meaningful correlation between the predicted and actual results. In contrast, the embeddings from the third layer of MatBERT yield the highest correlation (Correlation\_1=39.61\%) with experimental results and exhibits the weakest influence from the tokenizer (Correlation\_2=17.01\%). Detailed ranking by the third layer of MatBERT is available in Table S1 in supplementary).

The Fig.S1 in suuplementary shows that the majority of the 84 material names tokenized by MatBERT tokenizer exhibit token lengths predominantly falling within the range of 7 to 11 tokens. By exclusively retaining this subset of material names that possess more closely aligned token lengths (39 material names), we re-ranked them using the embeddings derived from the third layer of MatBERT and the correlation with experimental results achieved 51.34\%.

\subsection{GPT embedding and ChatGPT}
We employed the text-embedding-ada-002 model to generate embeddings for the 84 material names and subsequently ranked them based on their cosine similarity scores with center word. The results revealed a relatively lower correlation (23.55\%) between the predicted ranks and the actual experimental results (detailed ranking is available in Table S2 in supplementary). This performance metric fell notably short in comparison to the results achieved using context-average method and the MatBERT model. This further underscores the significance of domain-specific training and the provision of requisite context.

We leveraged ChatGPT to re-rank the provided dataset comprising 84 distinct materials (detailed ranking is available in Table S2 in supplementary). The model was tasked with considering various factors such as electrical and thermal conductivity, as well as the Seebeck coefficient, to assign novel rankings. The results demonstrated a moderate correlation of 24.03\% with the actual experimental results. This performance still fell short of the correlation achieved by the MatBERT model using context-average method, but slightly surpass that achieved through the utilization of GPT embeddings. We also tested few-shot prompting (using 1/5 ranked materials as few-shot and let ChatGPT rank rest 4/5 materials), and the averaged correlation of 5-fold is 30.73\%. Through an analysis of the outcomes generated by ChatGPT, we deduced that ChatGPT itself possesses a degree of domain-specific knowledge, enabling it to discern and recommend materials commonly utilized in specific applications from the provided list. However, its proficiency in comprehending less conventional or less prevalent materials appears to be limited.

\subsection{Fine-tuning effectiveness}
To evaluate the effectiveness of the finetuning process, we input material names (context-free) and context sentences including material names (context-average) to sentence embedding models SentMatBERT and SentMatBERT\_MNR fine-tuned on AllNLI, QQP and material description dataset. The spearman correlations can found in table \ref{tab:all}, which also includes the best results of word embedding extraction from BERT and MatBERT, and results of GPT embeddings and ChatGPT for comparison. 

\begin{table}[h]
\small
  \caption{\ All models evaluated with both methods. The evaluation metric for model performance is the Spearman correlation coefficient between predicted material ranking for certain application and the standard ranking, the former based on cosine similarity scores between material embeddings and the application embedding, the latter based on experimented results on zT scores. The comparison is made between token embedding models, pre-trained general purpose LLMs, and sentence embedding models. (ctx: context, zero: zero-shot learning, few: few-shot learning \textit{*Since OpenAI never officially announce the model parameter size, it's a rough estimation from their public technical report}})

\label{tab:all}
  \begin{tabular*}{0.48\textwidth}{@{\extracolsep{\fill}}llll}
    \hline
    Model & \makecell{Spearman \\ (ctx-free)} & \makecell{Spearman \\ (ctx-average)} & \makecell{Parameter \\ size} \\
    \hline
    BERT\_third\_layer & -0.0290 & 0.0846 & 108B \\
    MatBERT\_third\_layer & 0.1707 & 0.3961 & 108B \\
    OpenAI Embedding(ada v2) & 0.2355 & / & N/A* \\
    ChatGPT (GPT-3.5,zero) & 0.2403 & / & 1T-1.8T* \\
    ChatGPT (GPT-3.5,few) & 0.3073 & / & 1T-1.8T* \\
    SentMatBERT & -0.0084 & 0.2870 & 108B \\
    SentMatBERT\_MNR & 0.3145 & 0.5919 & 108B \\
    \hline
  \end{tabular*}
\end{table}

\textbf{Model differences} BERT and MatBERT share the same architecture, but while BERT was pre-trained with general texts, MatBERT incorporates material knowledge into its word embeddings. At the third layer, MatBERT with context-free inputs managed a weak but positive correlation 0.1707, which albeit unsatisfying is better than the result of original BERT -0.290, confirming the significance of domain-specific training. Judging by a spearman correlation very close to zero -0.0084, SentMatBERT does not recognize relations between the center word “thermoelectric” and context-free material names. With sentential inputs, the result of SentMatBERT (0.2870) is 11 points inferior than that of MatBERT\_third\_layer (0.3961). The poor outcomes of SentMatBERT indicate that simply pooling all word embeddings into one sentence embedding is severely inadequate.

\textbf{Method impact} The model inputs of context-free evaluation method are merely chemical formulae such as “Cu2Se”, which are composed of only a few tokens, therefore doesn’t fully realize the potential of our sentence embedding models. With context-average method, SentMatBERT and SentMatBERT\_MNR push the spearman correlation higher by approximately 28 points, and MatBERT\_third\_layer by 22 points. As the improvement suggests, giving a textual context to each material name allows models to capture semantic meanings more adeptly. Since transformers architecture relies on the attention mechanism, it is expected for transformer models to showcase augmented proficiency at downstream tasks when input with full sentences rather than isolated words. 

\textbf{Finetuning impact} SentMatBERT\_MNR, finetuned through Triplet and Siamese network, achieves an substantial improvement compared to SentMatBERT without any finetuning. For both context-free and context-average evaluation methods, the metric is increased by over 30 points due to finetuning. SentMatBERT\_MNR with contextual inputs yields a relatively strong correlation 0.5919 between model prediction and gold labels, surpassing MatBERT\_third\_layer by 20 points, the DFT baseline by 28 points, and Word2vec results by 7 points, which yields the best result at thermoelectrical material prediction task in this study. Therefore It is reasonable to conclude that our two-step finetuning approach to the end of producing semantically rich sentence embeddings manifests great efficacy. 

\section{Discussion}
In prior research \cite{tshitoyan2019unsupervised, venkatakrishnan2020knowledge}, there was a hypothesis that contextual embeddings would outperform traditional Word2Vec embeddings in material prediction task, and indeed, they have exhibited promise in some knowledge discovery tasks \cite{panesar2022biomedical, deb2022comparative}. However, our practical exploration in this study has unveiled nuanced findings. While contextual embeddings do possess the capability to predict material performance for a certain application to some extent (40\%) and surpasses the correlation obtained by Density Functional Theory (DFT) predictions (31\%) \cite{tshitoyan2019unsupervised}, they fall short if without enhancement when compared to conventional Word2Vec methods (52\%) \cite{tshitoyan2019unsupervised}. We adapted methods for enhancing BERT embeddings for sentence representation in NLP to the material prediction task, fine-tuning a SentMatBERT\_MNR on both a general sentence similarity task and our self-constructed material description similarity task. This model achieved a Spearman correlation of 0.5919 between model predictions and gold labels, surpassing the DFT baseline by 28 pointsand Word2Vec method by 7 points. In addition, contextual embeddings are more flexible than Word2Vec embeddings, as they do not suffer from the out-of-vocabulary problem and capture more experimental information hidden in scientific literature. In the future, we plan to conduct more experiments with other language models and further improve the effectiveness of contextual embeddings in material prediction.

In our experiments, the substantial disparity in performance observed between MatBERT and the BERT model on the same method underscore the indispensability of domain-specific training when employing contextual embedding models for material prediction tasks. In other downstream domain-specific NLP tasks, such as Named Entity Recognition (NER), the availability of labeled data (there is usually annotated data for supervised fine-tuning), coupled with the inherent capabilities of BERT models, enables the narrowing of the knowledge gap. Consequently, the disparity in performance between BERT and MatBERT in these downstream tasks does not tend to be exceedingly pronounced \cite{trewartha2022quantifying}. In the contrast, material prediction is an unsupervised task without annotated data and primarily rely on the model's capacity to encapsulate latent knowledge embedded within material names, highlighting greater importance of domain-specific pretraining. 

Through comparisons of different methods for obtaining contextual embeddings, we have found that utilizing the output embeddings from the third layer of contextual embedding models (BERT and MatBERT in our study), in conjunction with the context-average method (averaging embeddings of material names generated from various context sentences), is the most suitable approach for this specific task. Comparing the results across layers, the correlation values peak at the third layer and rapidly decline with a resurgence towards the end, which follows a similar trend in word similarity task \cite{bommasani2020interpreting}. Prior work has suggested that for most language models, the lower layers specialize in local syntactic relationships while the higher layers may be expressly encoding contextual semantic information \cite{peters2018dissecting, tenney2019bert, liu2019linguistic}. In synthesizing these pieces of information, it can be inferred that the characterization of material names primarily stems from their intrinsic lexical-level information. While necessitating a certain degree of contextual information, this reliance on context is less pronounced compared to more advanced semantic-level tasks such as coreference identification. This observation indirectly lends support to the notion that the straightforward application of Word2Vec yields good performance in this task.

Additionally, our experiments have shed light on the "tokenizer effect", where contextual embeddings tend to prioritize material names with shorter tokenized lengths when ranking materials. We quantified the strength of this tokenizer effect using Spearman correlation and observed that good prediction performance is often associated with low tokenizer effect but low tokenizer effect does not have to be related with good prediction performance. In other words, low tokenizer effect is a necessary but not sufficient condition for good prediction performance of contextual embeddings.

The results of our experiments, along with a series of tokenizer-related tests, suggest that standard tokenization methods like WordPiece may not be suitable for capturing the terminology in material science and further improvements in the effectiveness of contextual embeddings for material prediction necessitate refining the current tokenizer mechanisms. Although MatBERT tokenizer was pretrained on domain-specific text, it cannot guarantee that longer material names are not overly segmented, potentially leading to loss of meaning when aggregating token embeddings to form the overall material name embedding. One potential approach entails augmenting the vocabulary with as many complete material names as feasible, replacing subwords of material names, prior to domain-specific pretraining. The other approach is to use paragraph-level text as the source of representation of a material name, to bypass “tokenizer effect” mechanism. Comparing two approaches, the second one is operationally more feasible and has the ability to leverage contextual embedding for the encoding of higher-level information. The source text can be formed by different kinds of information, such as structure description, property data and experimental performance.

We also employed GPT embeddings and ChatGPT for the material prediction task. Intriguingly, ChatGPT exhibited a slightly better performance to GPT embeddings, autonomously providing limited insights without specific prompts. Their performance fell short of the MatBERT+context-average approach, but outperformed MatBERT+context-free approach. This suggests that there may be room for exploration in adopting a hybrid approach that leverages contextual information from sentences to enhance GPT's comprehension of materials, potentially yielding improvements in predictive accuracy.

\section{Conclusion}
To generate semantically rich representations of materials, we compared two-word embedding models, BERT and MatBERT, along with two sentence embedding models, SentMatBERT and SentMatBERT\_MNR. We present the sentence embedding model SentMatBERT\_MNR finetuned with natural language inference triplets and material description pairs in a contrastive learning framework for application-oriented material prediction tasks. Additionally, we propose the evaluation method context-average, where models are provided with context sentences encompassing chemical formulae, which results in enhanced model performance. SentMatBERT\_MNR with contextual inputs achieves a 0.5919 Spearman correlation between model prediction and gold labels, surpassing the baseline by 28 points. We also confirm that sentence embeddings derived from mean pooling of word embeddings do not ensure rich semantic meanings and that our two-step finetuning process is highly effective.

Our findings highlight the challenges and opportunities in applying advanced language processing techniques to material science. In this study, we demonstrate the application of LMs embeddings to thermoelectric material datasets, anticipating broader applications across various materials science domains. However, we acknowledge potential biases in the scientific literature used to generate context sentences, which may limit the model's ability to generalize to less-explored or emerging materials. We propose exploring domain-specific data augmentation techniques, such as using larger parameter LMs (e.g., GPT-4 or Claude 3.5) to generate material descriptions and mitigate biases from the context-average approach. Future work will also focus on incorporating more diverse and extensive material datasets for evaluation. From an algorithmic perspective, developing specialized tokenization methods for chemical formulas and material-specific terminologies could significantly improve the model's understanding of complex chemical structures. Furthermore, integrating contextual embeddings with advanced multi-modal models that combine both text and structural information may enhance material property predictions and accelerate application-oriented tasks. These improvements aim to address current limitations, push the boundaries of LMs applications in materials science, and enable more accurate predictions across diverse materials and properties.

In summary, by bridging the gap between natural language processing and materials science, this study accelerates material discovery and offers a template for applying similar techniques in other specialized scientific domains. As we refine these methods, integrating contextual embeddings with other advanced models may further enhance our ability to extract valuable insights from scientific literature, potentially revolutionizing the field of material science and beyond.

\section*{Author Contributions}
Yuwei Wan contributed to data curation, formal analysis, methodology, resources, and writing the original draft. Tong Xie was involved in conceptualization, data curation, formal analysis, funding acquisition, methodology, resources, and writing the original draft. Nan Wu contributed to data curation, methodology, visualization and writing the original draft. Bram Hoex, Wenjie Zhang, and Chun Yu Kit all participated in writing - review \& editing, and visualization. All authors have read and agreed to the published version of the manuscript.

\section*{Conflicts of interest}
There are no conflicts to declare.

\section*{Acknowledgements}
We gratefully acknowledge the support provided by Microsoft Research through the Accelerate Foundational Model Research project and the computational resources and support from NCI Australia and the Pawsey Supercomputer Centre, all of which have been instrumental in advancing this work. Tong Xie is grateful for the PhD scholarship from the UNSW Materials \& Manufacturing Future Institute. The authors extend their sincere thanks to all individuals and institutions whose assistance has contributed to the successful completion of this study.

\balance


\bibliography{rsc} 
\bibliographystyle{rsc} 

\clearpage
\onecolumn

\end{document}